\documentclass[letterpaper, 10 pt, conference]{ieeeconf} 
\IEEEoverridecommandlockouts                         
\usepackage{color} 
\usepackage{graphicx} 
\usepackage{hyperref}
\usepackage{amsmath} 
\usepackage{booktabs} 
\usepackage{amsfonts}

\title{\bf
RAG-6DPose: Retrieval-Augmented 6D Pose Estimation \\ via Leveraging CAD as Knowledge Base
}

\author{Kuanning Wang$^{1}$, Yuqian Fu$^{2,\dagger}$, Tianyu Wang$^{1}$, Yanwei Fu$^{1}$, Longfei Liang$^{3}$, Yu-Gang Jiang$^{1}$, Xiangyang Xue$^{1}$
\thanks{
$^1$~Fudan University, China.
$^2$~INSAIT, Sofia University “St. Kliment Ohridski”, Bulgaria. 
$^3$~NeuhHelium Co.,Ltd., China. }
\thanks{$\dagger$ indicates the corresponding author: Yuqian Fu (yuqian.fu@insait.ai)}
}

\begin{document}

\maketitle
\thispagestyle{empty}
\pagestyle{empty}

\begin{abstract}
Accurate 6D pose estimation is key for robotic manipulation, enabling precise object localization for tasks like grasping. We present \textit{RAG-6DPose}, a retrieval-augmented approach that leverages 3D CAD models as a knowledge base by integrating both visual and geometric cues. Our RAG-6DPose roughly contains three stages: 
1) Building a Multi-Modal CAD Knowledge Base by extracting 2D visual features from multi-view CAD rendered images and also attaching 3D points;
2) Retrieving relevant CAD features from the knowledge base based on the current query image via our ReSPC module; and 3) Incorporating retrieved CAD information to refine pose predictions via retrieval-augmented decoding. Experimental results on standard benchmarks and real-world robotic tasks demonstrate the effectiveness and robustness of our approach, particularly in handling occlusions and novel viewpoints.
Supplementary material is available on our project website: \href{https://sressers.github.io/RAG-6DPose}{https://sressers.github.io/RAG-6DPose} .

\end{abstract}

\section{Introduction}
Monocular 6D pose estimation aims to accurately predict an object's 3D position and orientation from a single RGB image, making it crucial for tasks such as robotic grasping and interaction. However, achieving robust 6D pose estimation remains challenging due to factors such as occlusions (including self-occlusions), lack of object textures, and the domain gap between synthetic and real-world data.

\begin{figure}
  \centering
  \includegraphics[width=0.95\linewidth]{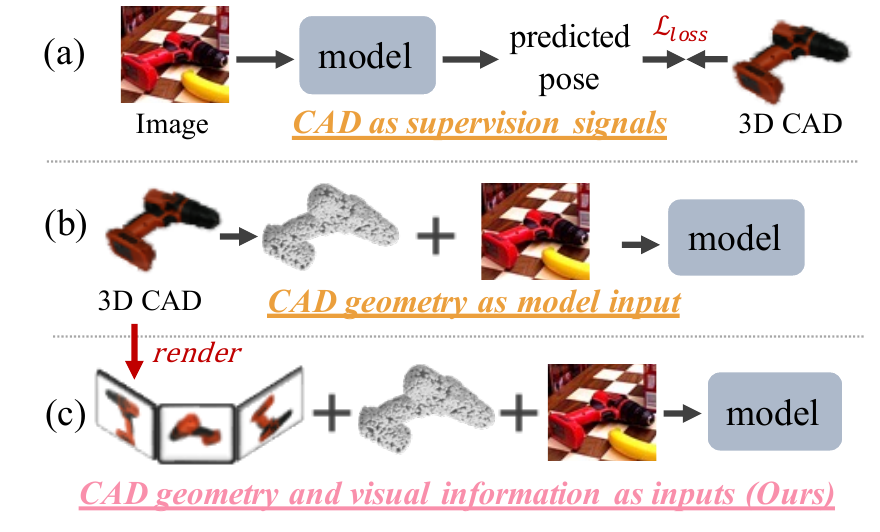}
   \vspace{-0.2in}
  \caption{Comparison of methods: Prior works (a) and (b) use CAD for solely supervision or geometry input, while ours (c) explores both CAD geometry and visual appearance for enhanced pose estimation.
   \vspace{-0.2in}
   \label{fig:teaser} }
  \vspace{-0.1in}
\end{figure}

Given one query RGB image and its corresponding 3D CAD model, many methods~\cite{Xiang2017PoseCNNAC, Li2024MRCNet6P, Li2019CDPNCD} obtain the object pose using the image as input while treating the CAD model solely as a supervision signal. 
We believe the valuable information, e.g., spatial relationships and visual appearance in CAD, should be actively explored. 
Some works~\cite{Li2022DCLNetDC, Jiang2023CenterBasedDP} with depth input have attempted to take CAD as direct input to the model, bringing geometric information into the learning process via techniques like point-based encoding, etc.
Inspired by those works but going beyond them, we propose to explore the full usage of CAD, not only by leveraging its geometric properties but also by incorporating the often-overlooked appearance as input to the model.
A summary of CAD utilization approaches is shown in Fig.~\ref{fig:teaser}, highlighting our early exploration of \textit{integrating visual and geometric information into the model for pose estimation}.

Technically, achieving this integration poses two key challenges: 1) \textit{Cross-modal discrepancy:} Visual feature extraction is typically conducted in 2D, while geometric information (e.g., coordinates) naturally resides in 3D, leading to a cross-modal gap. %Additionally, the query image belongs to a different modality than the 3D CAD model, further complicating alignment. 
2) \textit{Efficient information retrieval:} Not all CAD features are equally relevant—only the regions corresponding to the query image contribute effectively to pose estimation. Thus, how to dynamically retrieve the most useful features for the query image is also crucial.

To address these challenges, we introduce \textbf{RAG-6DPose}, a retrieval-augmented pose estimation method that leverages CAD as the knowledge base. 
RAG-6DPose follows a structured three-stage pipeline: 
a) \textbf{Building a Multi-Modal Knowledge Base:} 
Given that current models, e.g., DINOv2~\cite{Oquab2023DINOv2LR}, excel at extracting 2D features compared to 3D, we represent CAD visual features in 2D while preserving their 3D geometric properties. Specifically, we use DINOv2 to extract visual features from multi-view images rendered from CAD. These features are then mapped back to 3D points using depth. Finally, each point integrates visual features, coordinates, and color, forming a rich multimodal CAD knowledge base.
b) \textbf{Retrieving CAD Information:} 
This step involves retrieving relevant features from the knowledge base based on the RGB input. To accomplish this, we introduce the ReSPC module, which performs a retrieval operation that facilitates the extraction and fusion of both geometric and visual appearance information, thereby effectively retrieving the CAD information that best corresponds to the input image.
c) \textbf{Incorporating Retrieved CAD for Pose Estimation:} Finally, we integrate the retrieved CAD features into the final output through retrieval-augmented decoding.

Extensive results show state-of-the-art (SOTA) performance, confirming the importance of integrating CAD’s visual and geometric information, especially the visual aspect. Real-world robotic experiments further demonstrate the application of our method in the case of object grasping.

\section{Related Work}
\noindent \textbf{6D Pose Estimation.} 
Current pose estimation approaches could be broadly categorized into direct regression-based methods and 2D-3D correspondence-based methods, each with distinct core concepts and approaches.
Direct regression-based methods use end-to-end models to directly regress 6D pose parameters.
For instance, PoseCNN~\cite{Xiang2017PoseCNNAC} uses a CNN to regress 3D translation and rotation from the input RGB image. Other notable methods include PoseNet~\cite{Kendall2015PoseNetAC}, GDRNet~\cite{Wang2021GDRNetGD}, and MRCNet~\cite{Li2024MRCNet6P}. Typically, these methods involve a single stage that extracts image features and predicts 6D poses, though some, like MRCNet, incorporate an additional stage for refined pose estimation through multi-scale residual correlation.

Different from the direct regression-based methods, 2D-3D correspondence-based methods~\cite{Peng2018PVNetPV, Park2019Pix2PosePC,Haugaard2021SurfEmbDA} propose to establish correspondences between 2D image pixels and 3D model points, solving pose with algorithms like PnP-RANSAC~\cite{fischler1981random}.
This type could be further categorized into sparse and dense correspondence approaches:
Sparse methods, such as PVNet~\cite{Peng2018PVNetPV}, use keypoints to link 2D and 3D features and are efficient but sensitive to keypoint detection. 
Flagship examples of dense methods include Pix2Pose~\cite{Park2019Pix2PosePC}, SurfEmb~\cite{Haugaard2021SurfEmbDA}, DPODv2~\cite{Shugurov2021DPODv2DC}.
Typically, SurfEmb~\cite{Haugaard2021SurfEmbDA} establishes dense correspondences between 2D pixels and 3D CAD points in a self-supervised fashion, which achieves significant improvements over previous supervised 2D-3D correspondence methods.

According to how CAD is being used, we can also categorize those methods into: a) those use CAD solely for supervision, most of the regression-based methods~\cite{Xiang2017PoseCNNAC, Wang2021GDRNetGD} fall in this group; b) those explicitly incorporate CAD geometry as input to learn the spatial geometric relationships, examples include well-designed 2D-3D correspondence methods~\cite{Haugaard2021SurfEmbDA}, 3D-3D correspondence~\cite{Li2022DCLNetDC, Gan2024PriorinformationguidedCP}, etc.
Compared with the prior works, our RAG-6DPose features integrate both CAD geometry and visual appearance, which aligns more closely with human intuitive visual perception.

\noindent \textbf{Retrieval Augmented Generation.} 
Retrieval Augmented Generation (RAG)~\cite{Lewis2020RetrievalAugmentedGF} has been shown to enhance the performance of large language models (LLMs), particularly in knowledge-intensive tasks, by mitigating issues such as opaque reasoning.
In this framework, a pre-trained model serves as parametric memory, while external non-parametric memory, e.g., a dense vector index of Wikipedia, provides searchable knowledge, achieving state-of-the-art performance over traditional models.
RAG techniques are now expanding beyond text~\cite{zheng2025retrieval}.
%For example, RA-CLIP~\cite{Xie2023RACLIPRA} enhances contrastive language-image training by retrieving relevant image-text pairs, and retrieval-based conditioning with DINOv2 features improves cross-modal generalization in medical segmentation~\cite{Zhao2024RetrievalaugmentedFM}.
For example, RA-CLIP~\cite{Xie2023RACLIPRA} enhances contrastive language-image training by retrieving relevant image-text pairs, RealRAG~\cite{lyu2025realrag} and Domain-RAG~\cite{li2025domain} leverage retrieved images to improve the text-to-image generation result.
Applications are also extending into multimodal industrial tasks~\cite{Riedler2024BeyondTO}, showing RAG’s versatility beyond conventional NLP.  In this paper, we explore an efficient retrieval-augmented method in 6D pose estimation, particularly for leveraging 3D CAD models.

\section{Methodology}

\begin{figure*}[t]
  \centering
  \includegraphics[width=1.\linewidth]{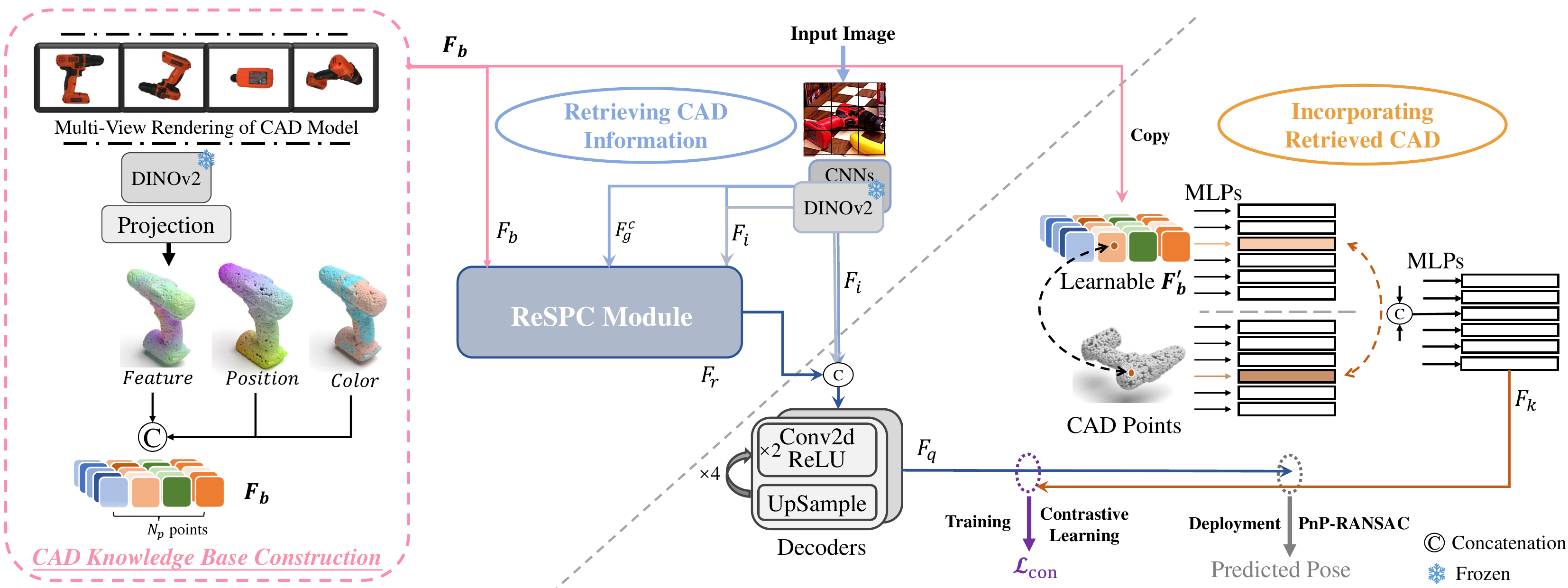}
  \vspace{-0.1in}
  \caption{\textbf{Method Architecture}. 
  The left part provides an overview of ``Building a Multi-Modal CAD Knowledge Base.''
  Particularly, we have two processes, ``Retrieving CAD Information'' and ``Incorporating Retrieved CAD for Pose Estimation'' to leverage the knowledge base \( F_b\). The diagonal dashed line separates the two processes, highlighting their distinct functions in our method. 
  \label{fig:fig2} }
  \vspace{-0.15in}
\end{figure*}

\noindent\textbf{Task Formulation.}
Given an RGB image \( I \) and a 3D CAD model \( M \) of a target object, 6D pose estimation requires the method to estimate the pose \( p \) of the object instance in the image \( I \) using the CAD model \( M \).

\subsection{Framework Overview}
\noindent\textbf{Preliminary.} Considering 
the prior SOTA performance achieved by SurfEmb~\cite{Haugaard2021SurfEmbDA}, a typical 2D-3D correspondence-based pose estimator, we establish our method following a similar pipeline with SurfEmb.  
Specifically, SurfEmb contains an \textit{encoder-decoder architecture} to extract the feature from image (\textit{query}, $F_{q}$), and a \textit{Siren MLP module}~\cite{Sitzmann2020ImplicitNR} to extract geometry i.e., point cloud features from CAD (\textit{key}, $F_{k}$). During the training stage, the \textit{query image feature $F_{q}$} and the \textit{key CAD geometry feature $F_{k}$} are used to perform the contrastive learning -- maximizing the similarity between corresponding 2D pixels and 3D points. When it comes to deployment, the $F_{q}$, $F_{k}$ are used to predict the final pose. 

We inherit the basic modules and contrastive learning from SurfEmb, while making several clear improvements: 1) We construct a rich multimodal CAD knowledge base and incorporate it into the vanilla encoder-decoder architecture via the idea of retrieval augmentation; 2) We also combine the 3D CAD multimodal feature with the CAD points making the final generated key CAD feature more comprehensive; 3) Besides, instead of using only CNN for extracting key image feature, we also introduce the DINOv2 into the encoder.

\noindent\textbf{Overall of Our RAG-6DPose.} 
The illustration of our method is provided in Fig.~\ref{fig:fig2}, including mainly three stages.
1) \textbf{Building a Multi-Modal CAD Knowledge Base:} 
Given the CAD model $M$, this step aims to extract both the visual and geometry information from $M$. Typically, the visual information is obtained in 2D space via extracting features from multiple rendered images, i.e., offline onboarding, while the geometry information is conveyed in 3D, i.e., the coordinate and color to each point. This step results in the multimodal CAD knowledge base $F_b$.
2) \textbf{Retrieving CAD Information:} 
This step takes a cropped image as input based on 2D object detection, using encoders (CNNs and DINOv2) to generate features $F_g^c$, $F_i$. The $F_g^c$ denotes the feature generated by one CNN solely, while $F_i$ denotes the combined feature from CNN and DINOv2.
Then our proposed retrieval module, namely ReSPC, takes the 2D image features $F_g^c$, $F_i$, and the CAD knowledge base $F_b$ as input, generating the retrieved CAD features $F_r$.
3) \textbf{Incorporating Retrieved CAD for Pose Estimation}: 
This step decodes the concatenated \( F_r \) and \( F_i \) generating the decoded \textit{query feature} \( F_q \). The \textit{key feature} $F_{k}$ is obtained by incorporating and embedding the multimodal CAD knowledge feature $F_b$ and CAD points.
With the query image feature $F_{q}$ and key CAD feature $F_{k}$ generated, following SurfEmb, contrastive learning is performed during training, resulting in \( \mathcal{L}_{con} \), while the deployment achieves pose prediction.

\subsection{Offline Multi-Modal Knowledge Base Construction}
To efficiently extract visual and geometric information from the given CAD $M$, we propose tackling them differently.
Our approach, shown in Fig.~\ref{fig:fig2}, rather than directly extracting features from the CAD model, we first render it from multiple viewpoints forming multi-view RGBD images \( \{ I_i \}_{i=1}^m \), we then adopt DINOv2~\cite{Oquab2023DINOv2LR} to extract visual appearance features for each view of that specific object forming  \( F_v^{(i)} \in \mathbb{R}^{H_v \times W_v \times D_v} \), where \(H_v\), \(W_v\) and \(D_v\) are the height and width of the feature map, respectively, and \( D_v \) is the feature dimension.
Each feature map is then upsampled via interpolation to match the original image resolution:
\[
F_{v}^{(i)}\rightarrow\widetilde{F}_{v}^{(i)}.
\]
After that, these multi-view image features are mapped back to the 3D CAD space with the rendered depth map.  This reprojection from 2D to 3D is inspired by recent methods~\cite{Wang2023D3FieldsD3, Caraffa2023FreeZeTZ,Huang2023VoxPoserC3}.
To align the rendered views and 3D CAD, we convert the rendered depth map \( I_d^{(i)} \) of each view into a point cloud \( P_d^{(i)} \), with each point corresponding to a pixel at 2D space.
For each sampled point \( p_k \) in the CAD model’s 3D point cloud with \( N_p \) points, we identify the nearest point \( p_{d_j} \) in \( P_d^{(i)} \):

$
p_{d_j} = \arg \min_{p_d \in P_d^{(i)}} \| p_k - p_d \|,
$

allowing the feature vector of each pixel in \( \widetilde{F}_{v}^{(i)} \) to be assigned directly to the corresponding CAD point.
To ensure comprehensive coverage, we sample from multiple viewpoints $\widetilde{F}_{v}^{(i)}$ and average the feature vectors for each CAD point across all views resulting in a visual feature set \( F_p = \{ F_p(p_k) \}_{k=1}^{N_p}\).

To finally integrate the 2D visual and the 3D geometry, for each point \( p_k \), we concatenate its positions, colors, and its visual feature. Note that the 3D coordinates are represented by 3D positional encoding. This leads to the multimodal knowledge base $F_b$.

\subsection{Retrieving CAD Information}
\begin{figure}[t]
  \centering
  \includegraphics[width=1.\linewidth]{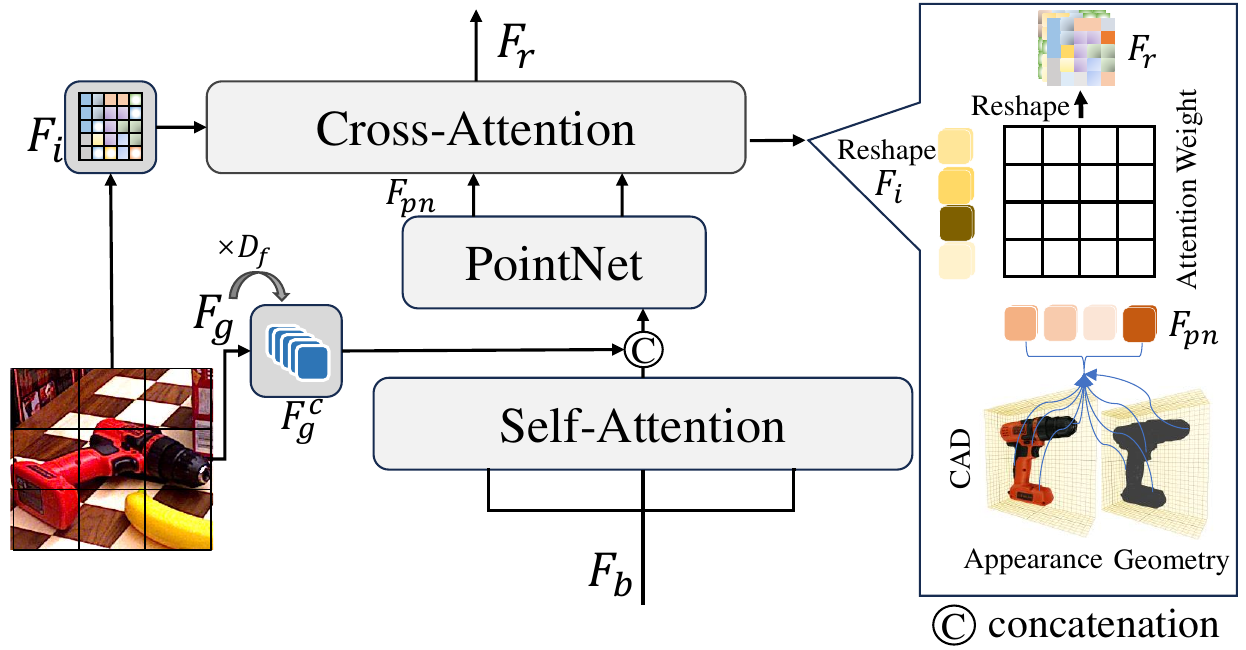} % retrieval.pdf
  \vspace{-0.1in}
  \caption{\textbf{ReSPC Module Architecture}. This module takes the CAD knowledge base \( F_b \) and image features as input and outputs the retrieved features \( F_r \). 
  The left part illustrates how we use encoders to extract image features. 
  The middle part demonstrates how attention mechanisms and PointNet process \( F_b \). 
  The right part details the function of Cross-Attention, illustrating the core retrieval process from the feature $F_{pn}$  extracted from the appearance and geometry of the CAD model.
  \label{fig:fig3} }
  \vspace{-0.15in}
\end{figure}

Given the multimodal CAD knowledge base $F_b$ and the input image $I$, the retrieval step is mainly achieved by our ReSPC module. As in Fig.~\ref{fig:fig3}, the ReSPC module is composed by a Self-attention, PointNet~\cite{Qi2016PointNetDL} and Cross-attention with the corresponding details as follow,

\noindent \textbf{Self-Attention.}
We feed the specific object CAD knowledge base \( F_b \) into our designed Multi-head Self-Attention module. 
It computes relationships among all feature elements, capturing both global context and local details.
To further process the data, we apply a linear layer in the attention module as, 
\[
   F_{sa} = SelfAttn(F_b),
\]
to reduce point dimensionality for efficient computation while preserving key information.

\noindent \textbf{PointNet.}
We extract global appearance features \( F_g \in \mathbb{R}^{1 \times D_g} \) with dimension \( D_g \) from the query image \( I \) using ResNet34 and replicate these features to form \( F_g^{\text{c}} \), 
which we treat as guidance for PointNet to process the multi-modal knowledge base effectively. 
We then concatenate \( F_g^{\text{c}} \) with \( F_{sa} \) and feed them into our PointNet~\cite{Qi2016PointNetDL}, which consists of multiple layers of Conv1d, normalization, and activation functions,
\[
F_{pn} = \text{PointNet}(\left[ F_{sa}, F_g^{\text{c}} \right]).
\]
It helps extract geometric information from the knowledge base, while also capturing local dependencies in the input feature sequence and ensuring more stable training through normalization. Through Self-Attention and PointNet, \( F_{pn} \) is enriched with fine-grained features capturing both the appearance and geometry of the CAD model.

\noindent \textbf{Cross-Attention.}
To extract features from RGB images, we use both DINOv2~\cite{Oquab2023DINOv2LR} and ResNeXt~\cite{Xie2016AggregatedRT}. The output features from both encoders are then concatenated and used as \( F_i \).
During training, we freeze DINOv2's parameters to save on computing resources. 
Both the construction of the knowledge base and the extraction of image features utilize DINOv2. Therefore, they naturally share consistent feature representations, facilitating more effective information interaction.
To efficiently retrieve the CAD knowledge base using image features, we introduce a multihead cross-attention module, 
\[
   F_{r} = CrossAttn(F_i, F_{pn}, F_{pn}),
\]
where \( F_i \) serves as the query, and \( F_{pn} \) acts as the key and value for the attention mechanism. The process is shown in Fig.~\ref{fig:fig3}.
This module measures similarity between \( F_i \) and \( F_{pn} \) through attention. Additionally, it enhances information interaction by incorporating geometric information in \( F_{pn} \) into the retrieval process, leading to more accurate and robust matching.
With DINOv2’s robust features for both the query image and CAD knowledge base, \( F_{r} \) captures the effective visual appearance and geometric attributes of CAD models related to the query image through our ReSPC module.

\subsection{Incorporating Retrieved CAD for Pose Estimation}

\noindent \textbf{Query Feature Extraction.}
The retrieved features \( F_r \), obtained through the ReSPC module, and together with the image features \( F_i \) are concatenated and fed into both decoders.
As shown in Figure~\ref{fig:fig2}, we use two decoders: one to decode the query features \( F_q \), structured as a U-Net~\cite{Ronneberger2015UNetCN}, and the other to decode the mask probabilities.

\noindent \textbf{Key Features Extraction.}
This module extracts key features \( F_k \) from the CAD model \( M \) in point cloud form and multi-modal knowledge base.
Given the learnable features \( F_b' \) copied from the CAD knowledge base (with \( F_b \) fixed), for each coordinate in the pose-specific rendered coordinates image or each 3D point \( p_j \) in \( M \), we locate the nearest point among the corresponding 3D coordinates in \( F_b' \) (based on 3D spatial distance) and assign it the associated feature vector \( f_j \) from \( F_b' \).
Using Siren~\cite{Sitzmann2020ImplicitNR} layers \( S_g \), \( S_v \), and \( S_i \), we then compute key feature as follows:\\
\[
f_k = S_i(\text{concat}(S_g(p_j), S_v(f_j)).
\]
The collection of all such \( f_k \) forms \( F_k \).

\noindent \textbf{Training: Contrastive Learning.}
The query consists of the decoded features \( F_q \). During training, the key features \( F_k \), corresponding to the rendered visible object coordinates of the ground truth, are treated as positives, while those uniformly sampled from the surface of the object model are considered negatives. The mask probabilities by mask decoder are used to isolate pixels belonging to the target object within the input image, allowing these pixels to be sampled in the rendered visible object coordinates map and the decoded query features \( F_q \), thereby forming positive query-key pairs.
Finally, \( F_k \) is optimized in contrastive learning against the decoder output \( F_q \). This process improves the consistency between 2D and 3D features.

\noindent \textbf{Training: Loss Function}. In the mask decoder, the final layer outputs a single channel, and a Sigmoid function is applied to obtain pixel-wise segmentation results. We utilize L1 loss \( \mathcal{L}_m \) for the segmentation task. For the feature decoder output, we adopt InfoNCE~\cite{Oord2018RepresentationLW} loss \( \mathcal{L}_{con} \) for contrastive learning. The point \(p_i\) corresponds to pixel coordinates \(c_i\).
\[
\mathcal{L}_{con} = -\log \frac{\exp(q_i k^+_i) }{\exp(q_i k^+_i)+ \sum_{\substack{j=1, j \neq i}}^N \exp(q_i k^-_j) },
\]\\
where $q_i = F_q(c_i)$, $k^+_i = F_k(p_i)$, $k^-_j = F_k(p_j)$, ${j\neq i}$.
The combined loss function is defined as:
\[
\mathcal{L} = \mathcal{L}_{con} + \alpha \cdot \mathcal{L}_m.
\]
where \( \alpha \) is a weighting factor to balance the contributions of segmentation and contrastive learning.

\noindent \textbf{Deployment: Pose Estimation and Refinement.}
During inference, we obtain the decoded query features \( F_q \) and generate key features \( F_k \) from the CAD. For pose estimation, query features are combined with masks and compared to key features to form a similarity matrix. 2D-3D correspondences are sampled based on similarity, and pose is estimated using PnP-RANSAC~\cite{lepetit2009ep,fischler1981random}. 
The training loss score helps evaluate and select the best pose hypothesis, which is then refined by maximizing the correspondence score to generate the final prediction.

\section{Experiments}

\subsection{Experimental Setup}
\noindent \textbf{Datasets.}
We evaluate the effectiveness of our method across a wide range of datasets, including the challenging LM-O~\cite{Brachmann2014Learning6O}, YCB-V~\cite{Xiang2017PoseCNNAC}, IC-BIN~\cite{Doumanoglou2015Recovering6O}, HB~\cite{Kaskman2019HomebrewedDBRD} and TUD-L~\cite{Hodan2018BOPBF} datasets.
These datasets feature multiple densely cluttered rigid objects with limited texture and significant occlusion, providing a comprehensive and rigorous evaluation that sufficiently validates the competitiveness of our approach.

\noindent \textbf{Model Details.}
We use the pre-trained DINOv2-Base and ResNeXt-101 32x8d as the backbone models. Our method is tested with both RGB-only input and RGB-D input, with the CAD model of the target object available in both cases.
For the detection of the 2D image, the default detection method of BOP~\cite{hodan2024bopchallenge2023detection} Challenge 2023 is used. 
In the Building a Multi-Modal CAD Knowledge Base stage, we follow the perspective selection strategy of CNOS~\cite{Nguyen2023CNOSAS}, choosing perspectives for each object and rendering corresponding images for each view.
During the encoder-decoder stage, the input image is first cropped around the object, resized to 224×224, and normalized.

\noindent \textbf{Training Details.}
The RAG-6DPose model is trained to convergence on the LM-O, TUD-L, IC-BIN, HB, and YCB-V datasets.
Synthetic images are used for LM-O, IC-BIN, HB, and YCB-V, while both synthetic and real images are used for TUD-L to evaluate the model's performance in mixed-domain scenarios.
We use the Adam optimizer with a fixed learning rate of 3e-5. 
Experiments are performed on RTX A6000 GPUs and RTX 3090 GPUs.

\noindent \textbf{Evaluation.}
For evaluation, we used the Average Recall metric, which is commonly used in the BOP Challenge. The Average Recall is the average of three indicators based on different error functions: VSD (Visible Surface Discrepancy), MSSD (Maximum Symmetry-aware Surface Distance), and MSPD (Maximum Symmetry-aware Projection Distance), as mentioned in work~\cite{Hodan2020BOPC2}.

\subsection{Comparison Results}
To comprehensively show the advantage of our proposed method, we compare our RAG-6DPose model with various prior state-of-the-art methods.
Note that for our model, we only train one model for all objects on one dataset, while some competitors, e.g., DPODv2~\cite{Shugurov2021DPODv2DC} and NCF~\cite{Huang2022NeuralCF} require training several different models for different objects. 

\begin{table}[h]
  \centering
  \small 
  \caption{\textbf{Comparison with state-of-the-art RGB based methods on BOP benchmarks.}  We report Average Recall in \% on \textbf{LM-O, IC-BIN, TUD-L, HB and YCB-V} datasets. P.E. means the number of pose estimators for an N-objects dataset. }
 \label{tab:1}
  \scalebox{.85}{
  \begin{tabular}{l|l|l|l|l|l|l} 
  \toprule
  % \hline
  \textbf{Method}           & LM-O & IC-BIN &  TUD-L & YCB-V & HB & \textbf{P.E.}  \\
  % \hline
  \midrule
    CDPNv2~\cite{Li2019CDPNCD} &62.4 & 47.3 & 77.2 & 53.2 & 72.2 & N \\
    DPODv2~\cite{Shugurov2021DPODv2DC} &58.4 & - & - & - & 72.5 &  N \\
    NCF~\cite{Huang2022NeuralCF}    &63.2 & - & - & 67.3 & - & N \\
    GDRNet~\cite{Wang2021GDRNetGD} &67.2 & - & - & - & - & N \\
    SurfEmb~\cite{Haugaard2021SurfEmbDA} &65.6 & 58.5 & 80.5 & 65.3 & 79.3 & N \\
    CosyPose~\cite{Labbe2020CosyPoseCM} &63.3 & 58.3 & 82.3 & 57.4 & 65.6 & 1 \\
    SO-Pose~\cite{Di2021SOPoseES} &61.3 & - & - & 66.4 & - & 1 \\
    CRT-6D~\cite{Castro2022CRT6DF6} &66.0 & 53.7 & 78.9 & - & 60.3 & 1 \\
    PFA~\cite{Hu2022PerspectiveFA}    &67.4 & - & - & 61.5 & 71.2 & 1 \\
    YOLO6DPose~\cite{Maji2024YOLO6DPoseEY} &62.9 & - &- & - & - & 1 \\
    MRCNet~\cite{Li2024MRCNet6P} &68.5 & - & - & 68.1 & -& 1 \\
    \hline
    \textbf{RAG-6DPose} & \textbf{70.0} & \textbf{60.1} & \textbf{83.3} & \textbf{68.6} &\textbf{85.3} & \textbf{1} \\
  \bottomrule
  \end{tabular}}
\end{table}

\noindent \textbf{Results on RGB based Pose Estimation.}
The main results are shown in Tab.~\ref{tab:1}. Overall, our method outperforms all the competitors across five datasets, with especially strong results on datasets with heavy occlusion like LM-O. 
Unlike approaches that require separate models for each object, RAG-6DPose solely needs one model, sharing parameters across all objects while retaining a small set of object-specific parameters. Despite this setup, it still outperforms others.
Compared to SurfEmb~\cite{Haugaard2021SurfEmbDA}, we achieve a 3.6\% improvement in average recall across 5 datasets, demonstrating a significant and consistent performance gain. Our method also outperforms the recent competitive state-of-the-art method MRCNet~\cite{Li2024MRCNet6P}, which uses the Render-and-Compare strategy, on both LM-O and YCB-V datasets, with a clear margin of 1.5\% on LM-O. This significant improvement validates the effective integration of the multi-modal CAD knowledge base in our model and highlights the effectiveness of the novel modules we introduced.

\noindent \textbf{Results on RGB-D based Pose Estimation.}  
We adopt the ICP~\cite{Rusinkiewicz2001EfficientVO} algorithm to refine pose estimation. 
With RGB-D input (see Tab.~\ref{tab:2}), our RAG-6DPose shows significant improvements over previous results (Tab.~\ref{tab:1}). We also outperform other RGB-based methods that use depth refinement.

\begin{table}[h]
  \centering
  \small 
  \caption{\textbf{Comparison with methods on BOP benchmarks based on RGBD and depth refinement.}  We report Average Recall in \% on \textbf{LM-O, IC-BIN and TUD-L} datasets.}
  \label{tab:2}
  \scalebox{.8}{
  \begin{tabular}{l|l|l|l|l} 
  \toprule
 \textbf{Method} & Domain & LM-O & IC-BIN &  TUD-L   \\
  \midrule
    Pix2Pose~\cite{Park2019Pix2PosePC} &RGB-D & 58.8 & 39.0 & 82.0  \\
    CosyPose~\cite{Labbe2020CosyPoseCM} & RGB-D& 71.4 & 64.7 & 93.9 \\
    SurfEmb~\cite{Haugaard2021SurfEmbDA} &RGB-D & 75.8 & 65.6 & 93.3 \\
    \hline
    \textbf{RAG-6DPose} & RGB-D & \textbf{76.8} & \textbf{68.7} & \textbf{93.9}  \\
  \bottomrule
  \end{tabular}}
\end{table}

\noindent \textbf{Comparison with 2D-3D Correspondence Methods.}
As shown in Tables \ref{tab:1} and \ref{tab:2}, our method outperforms 2D-3D correspondence-based approaches, such as DPOPv2~\cite{Shugurov2021DPODv2DC} and SurfEmb~\cite{Haugaard2021SurfEmbDA}. 
This improvement is due to our approach's ability to retrieve appearance features in CAD knowledge features. 
By leveraging this retrieval process, our approach effectively enhances the precision of pose estimation, especially in challenging scenarios with complex occlusions on the LM-O dataset.

\noindent \textbf{Results on RGB-based Methods For Each Object.}
To further investigate the performance of our method, we also analyzed the results for each object in the LM-O dataset. Specifically, we use $e$ as the pose estimation error and set the pose error thresholds $\theta_e$ to 5 and 10, which are the smallest two thresholds for fine-grained evaluation. A pose is considered correct if the calculated pose error $e$ satisfies $e \textless \theta_e$. We use $AR_{MSPD}^{e}$ to denote recall based on MSPD as a pose-error function. 
The final recall rate for each object is reported in Tab.~\ref{tab:4}. Comparing our RAG-6DPose with CDPNv2, SurfEmb, and MRCNet, we observe that our method achieves the best results on most objects.

\begin{table}[h]
  \centering
  \small 
  \caption{$AR_{MSPD}^{e}$ comparison results with RGB methods on BOP benchmarks using pose error threshold $\theta_e$=5 and $\theta_e$=10. 
  ``dri." means the ``driller" class, ``e.b." and ``h.p." denote ``eggbox" and ``helicopters" class respectively.
  }
  \label{tab:4}
  \scalebox{.78}{
  \begin{tabular}{l|l|l|l|l |l|l|l|l|l} 
  \toprule
 \textbf{Method}   & $\theta_e$ & \textbf{ape} & \textbf{can} & \textbf{cat} & \textbf{dri.} & \textbf{duck} & \textbf{e.b.} & \textbf{glue} & \textbf{h.p.}  \\

  \midrule
  CDPNv2   & 5    &  37       &  44     &  46     &  22             & 30          & 7           & 19      & 22  \\
    SurfEmb   & 5  &\textbf{52} &  46      & 50      &  28             & 36          & \textbf{27} &  23     & 19  \\
      MRCNet     & 5    &  44  & \textbf{56}&  53     &  40             & 39          & 22          & 25      & 22  \\
        \hline
        \textbf{RAG-6DPose}   &  5   &  51     & 52  & \textbf{53} & \textbf{41}     &\textbf{39}  & 27          & \textbf{25} & \textbf{26}  \\
        \hline
  CDPNv2   &  10   & 70       &  85     &  68     &  66             & 77          & 42          & 53      & 83 \\
  SurfEmb  &  10   &  77      &  92     & 78      &  82             & 78          & \textbf{70} & 74     & 89 \\
  MRCNet   &  10   &   77     & 92      &  81     &  82             & 82          & 64          & 74      & \textbf{90}\\
  \hline
  \textbf{RAG-6DPose} &  10 & \textbf{85}&\textbf{95}&\textbf{81}&\textbf{90}  &\textbf{82}  & 59          & \textbf{78}  & 87  \\
  
  \bottomrule
  \end{tabular}}
  \vspace{-0.1in}
\end{table}

\begin{figure}[t]
  \centering
  \includegraphics[width=.9\linewidth]{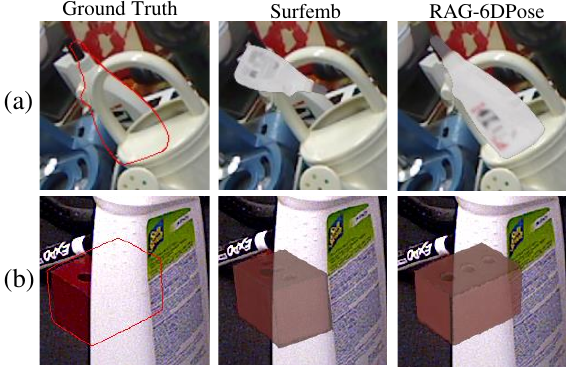}
  \caption{\textbf{Qualitative Comparison}. Parts (a) and (b) illustrate different scenes from the LM-O and YCB-V datasets. Each part displays three images (from left to right): the original RGB image (with the object's ground truth pose annotated), the SurfEmb result, and the RAG-6DPose result. This comparison underscores the effectiveness of RAG-6DPose in handling highly occluded scenes. \label{fig:fig4}}
\end{figure}
\subsection{Qualitative Results}
As shown in Fig.~\ref{fig:fig4}, we tested SurfEmb and our RAG-6DPose on challenging scenes with heavy occlusion from the LM-O and YCB-V datasets. In Fig.~\ref{fig:fig4}(a), SurfEmb struggles with complex occlusions involving both the foreground and the background, leading to poor pose estimation. 
In Fig.~\ref{fig:fig4}(b), SurfEmb shows some improvement when the object can be inferred from visible edges, but it still faces difficulties with severe occlusion. 
By contrast, RAG-6DPose delivers more accurate pose estimation and maintains robustness in complex environments.

\subsection{Ablation Studies}
As shown in Tab.~\ref{tab:5}, we conducted ablation studies on LM-O to evaluate the effectiveness of our proposed modules.
We categorize the experiments into full-scale and reduced-scale ablations. In the full-scale ablations, the model is trained on the entire dataset, while in the reduced-scale ablations, the model is trained on a subset comprising one-tenth of the full dataset until the error stabilizes.

\begin{table}[h]
  \centering
  \small 
  \caption{
  \textbf{Ablation Experiments}. Using RGB images from the LM-O dataset as input.  ``AR'' means average recall.}
  \label{tab:5}
  \scalebox{.82}{
  \begin{tabular}{l|c|c|c|c} 
  \toprule
    \textbf{Method}           & $AR\uparrow$ & $AR_{VSD}$ & 
  $AR_{MSSD}$ & $AR_{MSPD}$ \\
  \midrule
    \multicolumn{5}{c}{\textbf{Full-scale Ablation}} \\
  \midrule
  \textbf{RAG-6DPose}  & \textbf{70.0} & \textbf{53.4} & \textbf{69.1} & \textbf{87.3}\\
  $-$ 3D CAD Features & 66.5 & 50.1 & 64.8 & 84.7 \\
  $-$ DINOMLP & 68.8 & 52.5 & 67.7 & 86.4 \\
  C.A.Fusion $\rightarrow$ ConvFusion  & 68.6 & 52.1 & 67.7 & 86.0 \\ 
  C.A.Fusion $\rightarrow$ Avg & 67.7 & 51.4 & 66.5 & 85.2 \\
  ResNet34 $\rightarrow$ Avg &68.4  & 52.5 & 67.9 & 86.6 \\
  \midrule
    \multicolumn{5}{c}{\textbf{Reduced-scale Ablation}} \\
    \hline
    \textbf{RAG-6DPose}  & \textbf{59.3} & \textbf{43.2} & \textbf{55.5} & \textbf{79.3}\\
    PointNet $\rightarrow$ MLP  & 55.6 & 40.1 & 50.8 & 75.8 \\
    $-$ ResNeXt    & 47.4 & 29.9 & 39.0 & 73.2 \\
    $-$ \( F_p \)  & 57.8 & 42.0 & 53.6 & 77.9 \\
  \bottomrule
  \end{tabular}
  }
  \vspace{-0.1in}
\end{table}

\noindent \textbf{Settings.}
% Major
For full-scale ablations, we conduct 5 experiments, as shown in Tab.~\ref{tab:5}.
In the ``$-$ 3D CAD Features" (row 2), we removed the CAD knowledge base \( F_b \) and the ReSPC module.
In the ``$-$ DINOMLP" (row 3), we removed the learnable copy of the CAD knowledge base $F_b^{\prime}$. This leaves only the 3D coordinates processed by MLPs for key features, similar to SurfEmb, relying solely on geometric information. 
In the ``C.A. Fusion $\rightarrow$ ConvFusion" (row 4), we replaced the cross-attention with a convolution-based module, as in~\cite{Ren2022CorrI2PDI}, which simply fuses concatenated inputs without retrieval. 
In the ``C.A. Fusion $\rightarrow$ Avg''(row 5), we replaced the cross-attention with a simple fusion method that computes the average of \( F_i \) and \( F_{pn} \).
In the ``ResNet34 $\rightarrow$ Avg''(row 6), we replaced the global feature extracted by ResNet34 with the average of the ResNeXt feature.
% Minor
For reduced-scale ablations, we conduct 3 experiments.
In the ``PointNet $\rightarrow$ MLP,'' we replaced PointNet with an MLP.
In the ``$-$ ResNeXt'', we removed the ResNeXt.
Finally, in the ``$-$ \( F_p \)'', we removed the visual features \( F_p \) in knowledge base \( F_b \).

\noindent \textbf{Results.}
Tab.~\ref{tab:5} presents our results. The ablation experiments validate the effectiveness of our designs.
Notably, the ``$-$ 3D CAD Features'' shows a significant drop in the average recall, emphasizing the importance of retrieving detailed CAD visual and geometric information.
Both ``- DINOMLP'' and ``$-$ \( F_p \)'' demonstrate that integrating CAD visual appearance features enhances the model’s ability to learn robust 2D-3D correspondences.
Furthermore, ``C.A. Fusion $\rightarrow$ ConvFusion'', ``C.A. Fusion $\rightarrow$ Avg'', ``PointNet $\rightarrow$ MLP'', and ``ResNet34 $\rightarrow$ Avg'' show the effectiveness of the individual components within our ReSPC module.
Finally, ``$-$ ResNeXt'' demonstrates that incorporating the features retrieved using a frozen-parameter DINOv2 into the decoding process produces favorable results, which suggests that our retrieval method is effective.

\subsection{Robotic Experiment}
\begin{figure}[t]
  \centering
  \includegraphics[width=.9\linewidth]{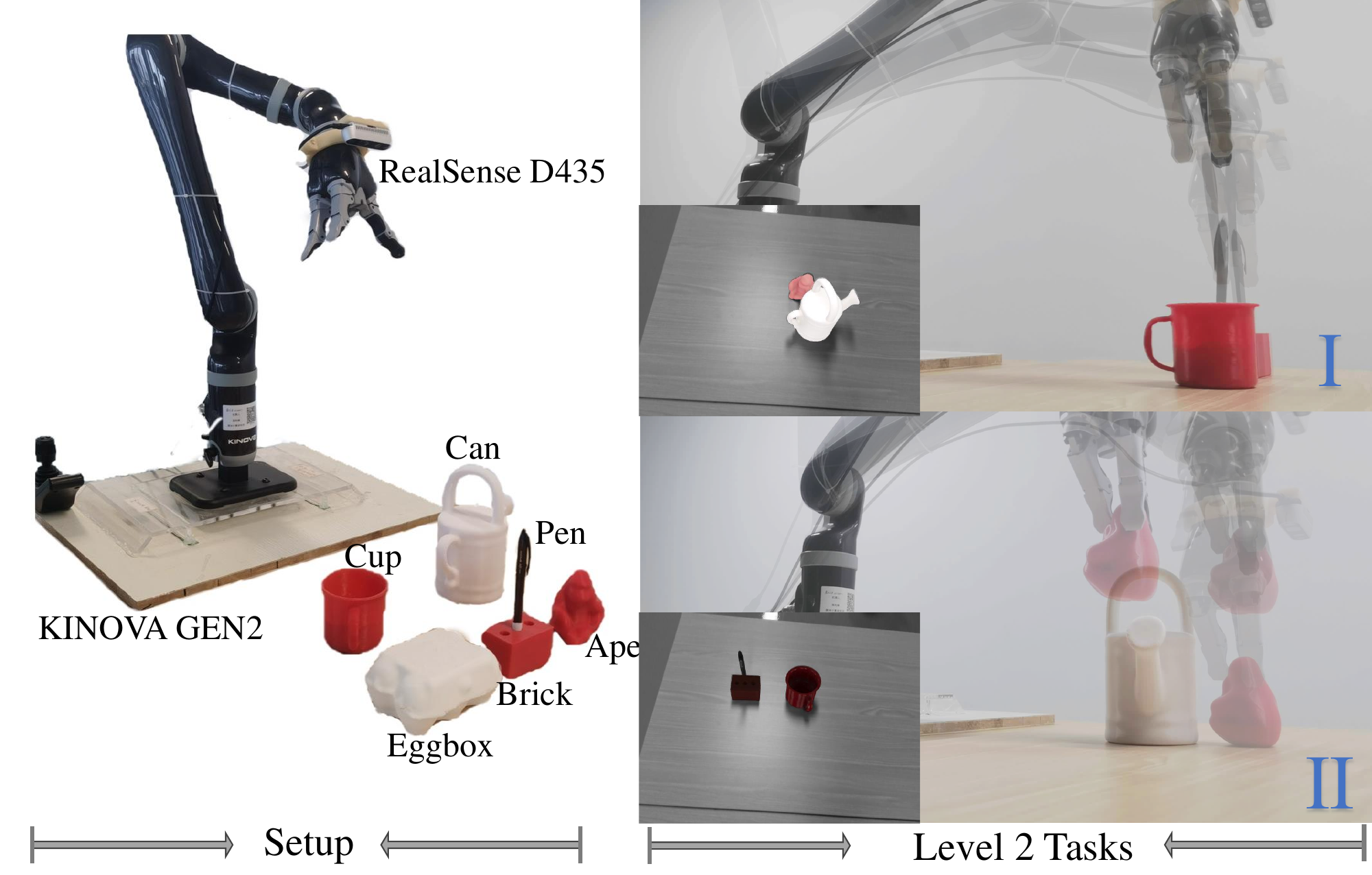}
%\vspace{-0.15in}
  \caption{
  \textbf{Robotic Experimental Setup.}
      \underline{\smash{Left}}: a third-person view of the setup. A \textbf{first}-person camera captures RGB image observations. We show different 3D printed objects from the LM-O and YCB-V datasets and a pen, all of which are used in our experiments.
      \underline{\smash{Right}}: we show ``Level 2'' (i.e., more challenging) tasks in our experiments. 
      We present the ``Pick and Place Pen'' task (I) and the ``Move Ape'' task (II).
      The bottom left corner shows the predicted object poses obtained from the observed image input. 
  \label{fig:realworld} 
  }
\end{figure}

\noindent \textbf{Setup.}
Our experimental setup is illustrated in Fig.~\ref{fig:realworld}. 
Our model is deployed on an NVIDIA RTX 6000 Ada Generation GPU to estimate the object's pose from RGB images captured by a RealSense D435 camera. We use a Kinova Gen2 robot arm equipped with three fingers. The experiments utilize objects from the LM-O and YCB-V datasets, inspired by home service and industrial applications.

\noindent \textbf{Task Description.}
We categorize our tasks into two difficulty levels. Level 1 tasks involve directly grasping specific objects on the table—namely, “Grasp Brick”, “Grasp Cup,” and ``Grasp Eggbox.'' Level 2 tasks are more challenging due to occlusions and the high precision required, and they necessitate the sequential estimation of two objects’ poses to complete the task, as illustrated in Fig.~\ref{fig:realworld}.
The first Level 2 task, named “Pick and Place Pen,” involves grasping a pen (not included in the datasets) inserted into a hole in the center of a brick and placing it into an adjacent cup, which requires accurate pose estimation for both the brick and the cup to guide the pen's pick-and-place actions.
The second task, called “Move Ape,” entails grasping an ape that is partially occluded by a watering can and moving it in front of the can’s spout, thus requiring precise pose estimation for both the ape and the watering can.

We use predefined grasp policies for the object model and, by combining them with the estimated object pose, transform these grasp poses into the robot frame for motion planning. 
Our evaluation metric focuses on task success rate. For Level 1 tasks, success is defined as a successful grasp without dropping the object. For Level 2 tasks, success is determined by whether the task is completed perfectly according to the specified requirements.
We repeat the experiment 10 times for each task. In repeated experiments, we randomize object placements and other aspects of the scene setup to increase the diversity of the settings.

\noindent \textbf{Results.}
The results for Level 1 and Level 2 tasks are shown in Tab.~\ref{tab:real}. Our approach achieves highly accurate pose estimation, particularly obtaining precise grasping poses for Level 2 tasks. The results demonstrate the effectiveness of our approach in real-world deployments.

\begin{table}[t]
  \centering
  \small 
  \caption{
  \textbf{Robotic Experiment Results}. We report the success rate of RAG-6DPose over 10 trials for each of 5 tasks.  
 }
 \label{tab:real}
  \scalebox{.78}{
  \begin{tabular}{c|c|c|c|c} 
  \toprule
  \multicolumn{3}{c|}{\textbf{Level 1}} & \multicolumn{2}{c}{\textbf{Level 2}}  \\
  \hline
   Grasp Brick & Grasp Cup & Grasp Eggbox &  Pick and Place Pen &  Move Ape  \\
  \midrule
  10/10 & 10/10 & 10/10 & 9/10 & 9/10 \\
  \bottomrule
  \end{tabular}
  }
  \vspace{-0.1in}
\vspace{-0.15in}
\end{table}

\section{Conclusion}
In this paper, we introduced a novel RAG-inspired approach, \textit{RAG-6DPose}, for 6D pose estimation from input RGB images. Our method aims to leverage the multimodal information in 3D CAD models, particularly the visual appearances, which have been less explored in prior works.
Technically, our approach consists of three key steps: building a multimodal CAD knowledge base, retrieving relevant CAD information, and incorporating the retrieved information into pose estimation. Extensive experiments, conducted with both RGB and RGB-D inputs, along with quantitative evaluations, ablation studies, visualizations, and detailed analysis, demonstrate the effectiveness and robustness of our method in tackling the challenging task of 6D pose estimation. Furthermore, real-world robotic experiments validate the successful application of our method for object grasping, further highlighting its practical utility.

%%%%%%%%%%%%%%%%%%%%%%%%%%%%%%%%%%%%%%%%%%%%%%%%%%%%%%%%%%%%%%%%%%%%%%%%%%%%%%%%
\bibliographystyle{IEEEtran}
\bibliography{main}

\end{document}